\documentclass[conference]{IEEEtran}
\IEEEoverridecommandlockouts

\usepackage{cite}
\usepackage{amsmath,amssymb,amsfonts}
\usepackage{algorithmic}
\usepackage{graphicx}
\usepackage{url}
\usepackage{textcomp}
\usepackage{xcolor}
\usepackage{hyperref}
\usepackage{multirow}
\usepackage{bm}
\usepackage{booktabs}
\usepackage{pifont}
\usepackage{makecell}
\def\BibTeX{{\rm B\kern-.05em{\sc i\kern-.025em b}\kern-.08em
    T\kern-.1667em\lower.7ex\hbox{E}\kern-.125emX}}
\begin{document}

\title{From Model Patterns to Abstract Semantics in Compositional Zero-Shot Learning
}

\author{
\IEEEauthorblockN{Weize Li\textsuperscript{1}
\quad Zhicheng Zhao\textsuperscript{1,2,3}$^\dagger$\thanks{$^\dagger$Correspondence Author: zhaozc@bupt.edu.cn.}
\quad Fei Su\textsuperscript{1,2,3}}
\IEEEauthorblockA{\textit{\textsuperscript{\rm 1}Beijing University of Posts and Telecommunications} \\
\textit{\textsuperscript{\rm 2}Beijing Key Laboratory of Network System and Network Culture}\\
\textit{\textsuperscript{\rm 3}Key Laboratory of Interactive Technology and Experience System} \\
\{bupt\_lwz, zhaozc, sufei\}@bupt.edu.cn}

}

\maketitle

\begin{abstract}
Compositional Zero Shot Learning aims to recognize unseen compositions by recombining learned primitives. Recent methods rely on vision language models and attempt to explicitly model contextual variations of primitives through multiple representations. However, such approaches are limited by fixed variant capacity and competition between abstract and concrete semantics. In this work, we present a new perspective that views primitive variations as the context-driven activation of concrete visual cues rather than independent entities. Based on it, we propose CLEAR, a CLoze-style rEAsoning-based Re-ranking framework inspired by human perceptual processes. CLEAR extracts conditional variants from the primitive candidate set in a coarse-to-fine manner, performs cloze-style reasoning to infer high-level semantics, and re-ranks predictions to correct biases toward salient concrete primitives. Extensive experiments demonstrate that CLEAR consistently improves the Base Model and outperforms state-of-the-art methods on the challenging C-GQA and MIT-States datasets. Code is available at \url{https://github.com/buptLwz/CLEAR}.

\end{abstract}

\begin{IEEEkeywords}
compositional zero-shot learning, reasoning, vision-language models, re-ranking
\end{IEEEkeywords}

\section{Introduction}
\label{sec:intro}
The ability to recognize unseen concepts by recombining previously acquired visual and semantic primitives constitutes a fundamental capability of the human brain. This process, termed compositional generalization, enables humans to efficiently acquire new knowledge. Inspired by it, Compositional Zero-Shot Learning (CZSL) has emerged as an important direction, aiming to endow models with similar generalization abilities. CZSL seeks to disentangle attributes and objects within compositions, so that novel compositions can be correctly recognized. For example, after learning a green bottle and a yellow pear, a model is expected to correctly recognize a green pear and a yellow bottle, even though this new composition have never been encountered during training.

\begin{figure}[htbp]
\centerline{\includegraphics[width=0.8\linewidth]{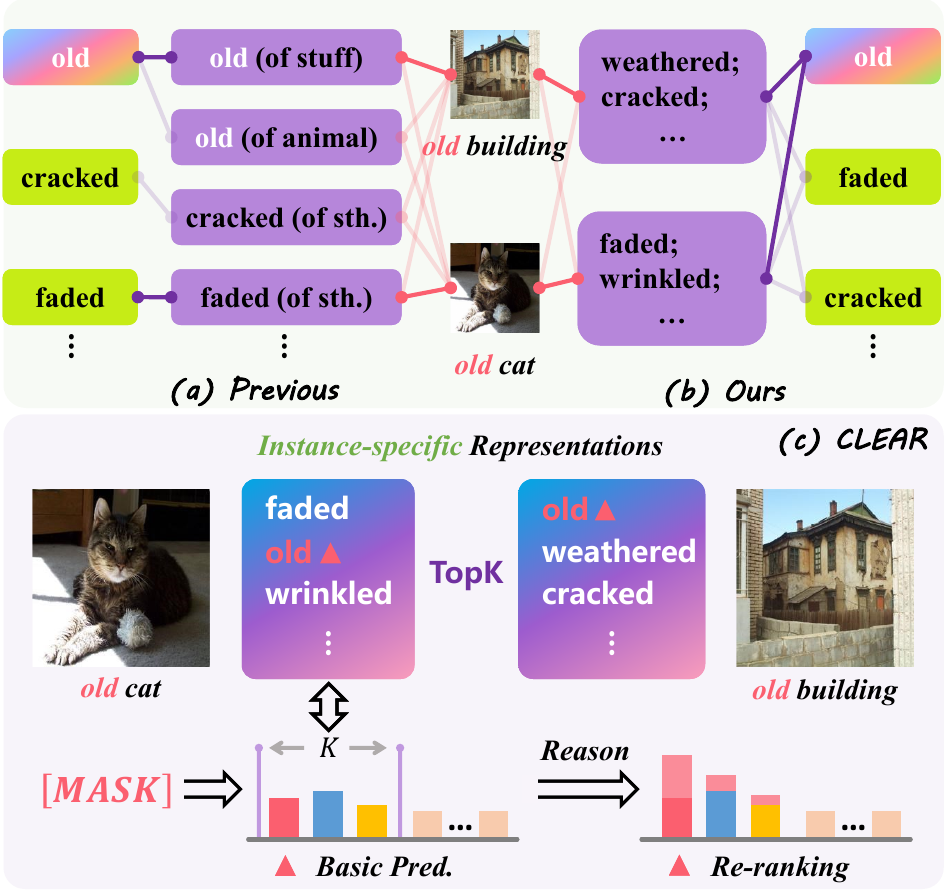}}
\caption{(a) Prior methods enumerate and model conditional variants for each primitive, but the number of variants is limited, and abstract and concrete semantics compete with each other;
(b) Our method first constructs a primitive candidate set and then reasons over it to infer high-level semantics. Built from primitive combinations, the candidate set not only has a higher information capacity but also avoids competition between semantics.
(c) The proposed CLEAR performs cloze-style reasoning over the candidate set derived from base predictions, and then re-ranks the base predictions to correct errors.}
\label{fig:vs}
\end{figure}
CZSL frameworks are mainly built upon pre-trained vision–language models (VLMs) such as CLIP\cite{clip}. By designing task-specific prompts and lightweight adaptations\cite{Troika}, existing methods aim to learn disentangled primitive textual representations. Although CLIP exhibits strong generalization capability, its performance on CZSL remains unsatisfactory. A primary reason lies in the strong contextual dependency of primitive semantics: the same primitive may exhibit substantially different visual manifestations across compositions, such as ``old cat" versus ``old building". Even within the same composition, different instances may emphasize different visual aspects of the same primitive, making it difficult to learn a single fixed representation that can simultaneously cover such variations.

Therefore, recent CZSL approaches attempt to explicitly model such contextual variations as shown in Fig.~\ref{fig:vs} (a), using techniques such as clustering\cite{CLUSPRO}, object-prioritized focus\cite{CPF}, or external structured knowledge\cite{logic}. They aim to enumerate contextual variants from the training data as comprehensively as possible, so that the same primitive can be recognized from multiple conditional perspectives.

However, rethinking this issue, we find that the contextual variation of primitives can not be the root cause, but rather a superficial manifestation. Many primitives in CZSL, especially abstract ones, are not atomic concepts, but emerge from the aggregation of multiple concrete visual cues. The conditional variants of a primitive can be viewed as emerging naturally from the context-driven activation of different visual cues. As shown in Fig.~\ref{fig:vs} (b), ``old" in ``old building" is inferred from cues such as ``weathered" and ``cracked", while ``old" in ``old cat" is inferred from cues such as ``faded" and ``wrinkled".

From this viewpoint, existing methods\cite{CLUSPRO,logic,CPF} essentially attempt to pre-model primitive variations by extracting commonly co-occurring visual patterns at the dataset level. While effective in capturing frequent contextual variations, such approaches inevitably face inherent limitations. First, they approximate primitive variations by a fixed and limited number of representations per primitive, failing to comprehensively cover long-tail unseen combinations or instance-specific fine-grained cues. Second, as shown in Fig.~\ref{fig:vs} (a), existing methods overlook an important fact: in the modern CZSL benchmarks, concrete primitives and abstract primitives typically coexist in the vocabulary and compete during prediction. Even if learned variants can summarize multiple concrete cues, highly salient visual details may still dominate the prediction and induce a bias toward concrete primitives.

To address these issues, inspired by the human perceptual process of progressively focusing when recognizing ambiguous concepts\cite{nature-human1,nature-human2}, we propose to model recognition as a coarse-to-fine procedure: instead of directly matching an exact answer in the full semantic space, the model first attend to multiple concrete cues and form a set of plausible, instance-specific hypotheses. Second, by reasoning over these hypotheses, it gradually infers the abstract concept that best explains the observed cues, alternatively, when the hypotheses are insufficient, it switches to selecting the most appropriate concrete primitive.

In brief, as shown in Fig.~\ref{fig:vs} (c), we propose a CLoze-style rEAsoning-based Re-ranking strategy (CLEAR). Specifically, we first employ a Base Model following a standard CZSL formulation, but instead of treating its predictions as the final output, we use the predicted candidate set to cover concrete primitives that are informative for inferring the abstract primitive. Then, CLEAR performs candidate-conditioned cloze-style reasoning in the textual space, mimicking the human process of summarizing abstract categories from plausible hypotheses. Finally, by re-comparing the reasoning results with primitive features, CLEAR re-ranks the Base Model’s predictions, thereby enabling reflective correction of the bias toward concrete primitives induced by salient visual details. Our contributions can be summarized as follows: 

1. We provide a new perspective on contextual variation in CZSL by arguing that primitive variants are not independent entities, but rather natural outcomes of context-driven activation of different visual cues.

2. We propose CLEAR, a coarse-to-fine CZSL framework that avoids explicit modeling of conditional structures and comprehensively covers semantic variations through instance-specific candidate sets.

3. CLEAR introduces a re-ranking strategy for CZSL, which effectively mitigates the bias toward concrete primitives induced by salient visual details.

4. CLEAR significantly improves the performance of the Base Model, outperforming SOTA methods on the challenging C-GQA and MIT-States datasets, while achieving competitive results on the simpler UT-Zappos dataset.
\begin{figure*}[!ht]
\centerline{\includegraphics[width=0.78\linewidth]{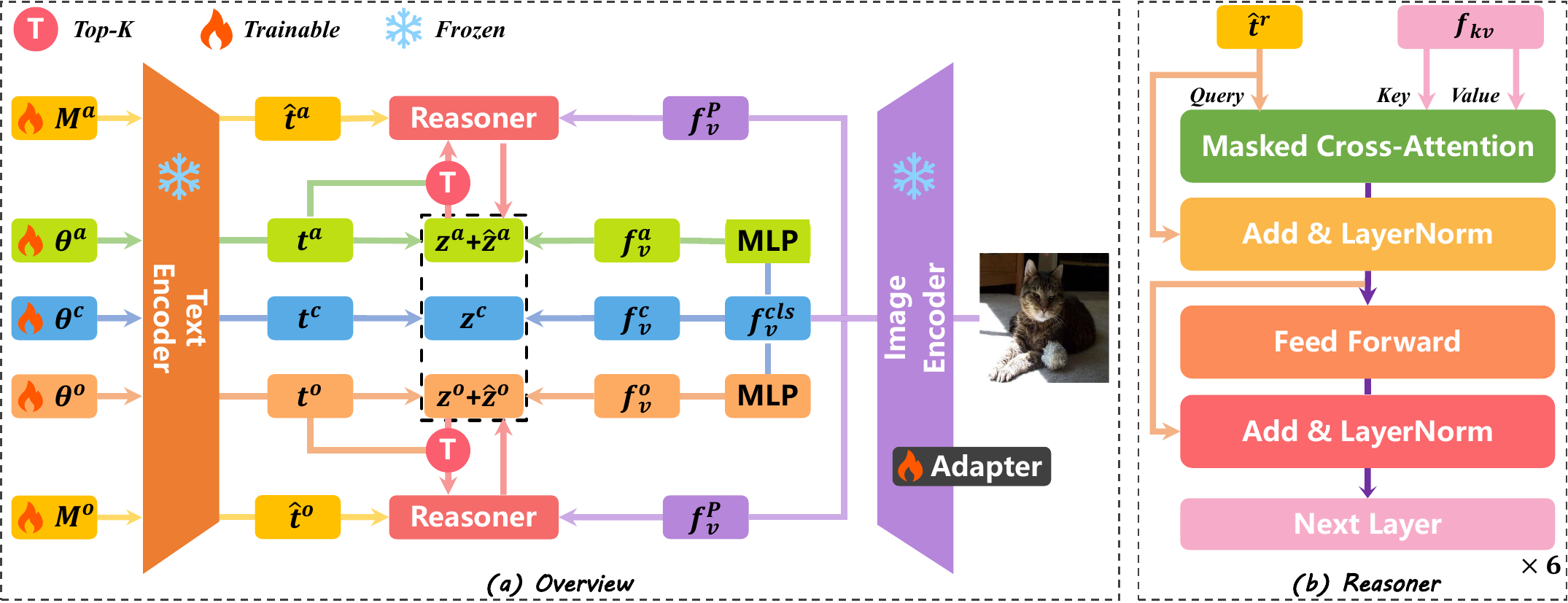}}
\caption{(a) Overview of CLEAR. The Base Model adopts a three-branch architecture to jointly fine-tune the CLIP visual encoder with adapters and learn CLIP-based textual primitive prompts $\theta^a$, $\theta^o$ and $\theta^c$. Based on the predictions of the Base Model, CLEAR uses learnable masked tokens $M^a$ and $M^o$ to decode high-level semantics from primitive candidates and visual patch features $f_v^P$. Finally, the re-ranking scores are computed by matching the high-level semantics and the primitive features in the textual space.
(b) The Reasoner consists of a set of lightweight Transformer decoders, which take masked representations as queries and the candidate set with visual patch features $f_v^P$ to serve as keys and values.}
\label{fig:overview}
\end{figure*}
\section{Related Work}
\subsection{Compositional Zero-Shot Learning}
Recently, VLMs have been increasingly applied to CZSL\cite{PLID,li2024context}, and the issue of conditional semantics has also garnered attention. CLUSPRO\cite{CLUSPRO} performs clustering in the visual space to enumerate contextual pattens and matches specific instances to prototypes. CPF\cite{CPF} prioritizes object localization, aggregating visual details across instances to form object-level posterior attribute representations. LOGICZSL\cite{logic} leverages LLMs to parse different primitives into external knowledge, explicitly training the conditional relationships. Other approaches\cite{VPCMJL,10737702} often place more emphasis on mitigating modality gaps and capturing visual details.


\subsection{Re-ranking}
Re-ranking has been extensively studied in retrieval systems\cite{rerankllm1,rerank1}, where it serves as a second-stage refinement to reorder a candidate list under a fixed semantic query. Such approaches focus on improving ranking accuracy by applying more expressive scoring functions, without altering the underlying semantic target. In CZSL, re-ranking acts as a reasoning mechanism that revisits hypotheses and reshapes primitive semantics, instead of merely reordering candidates under a fixed criterion.

\section{Method}
\subsection{Problem Formulation}

In CZSL, we define an attribute set $\mathcal{A} = \{a_1, a_2, \dots, a_M\}$ and an object set $\mathcal{O} = \{o_1, o_2, \dots, o_N\}$. The complete space of compositional concepts is given by their Cartesian product, i.e., $\mathcal{C} = \mathcal{A} \times \mathcal{O}$. We further consider an image set $\mathcal{X}$, where each image is annotated with a compositional label, and the set of image-associated compositions forms a subset $\mathcal{C}^x \subseteq \mathcal{C}$. The image set $\mathcal{X}$ and its corresponding composition set $\mathcal{C}^x$ are partitioned into two disjoint subsets: the seen compositions $\mathcal{X}_{s}$ with $\mathcal{C}^x_{s}$, and the unseen compositions $\mathcal{X}_{u}$ with $\mathcal{C}^x_{u}$, which defines the training set $\mathcal{T} = \{(x, c) \mid x \in \mathcal{X}^{s}, c \in \mathcal{C}^{s}\}$, and the test set $\mathcal{T}^{te} = \{(x, c) \mid x \in \mathcal{X}, c \in \mathcal{C}^x\}$, respectively. In the open-world setting, the training data and image collection remain unchanged, while the model is required to perform classification over the entire composition space $\mathcal{C}$ at test time.

\subsection{Overview of CLEAR}

Inspired by human behavior, CLEAR models the recognition process as a pipeline of coarse-grained classification, fine-grained reasoning, and re-ranking. Specifically, as illustrated in Fig.~\ref{fig:overview} (a), the final recognition results (the dashed box in the center) consist of two components: (1) the coarse-grained predictions for attributes, objects, and compositions, denoted as $\mathbf{z}^a$, $\mathbf{z}^o$, and $\mathbf{z}^c$, respectively; and (2) the re-ranking scores for attributes and objects, denoted as $\hat{\mathbf{z}}^a$ and $\hat{\mathbf{z}}^o$. Notably, the re-ranking is designed for primitives and does not include the compositional branch, where the conditional effects are explicitly constrained by a well-defined context. In CLEAR, the coarse-grained recognition is obtained by a Base Model through cross-modal matching between the visual features and the primitive feature sets. Based on the primitive candidates derived from the coarse-grained predictions and the patch-level visual features $f_v^P$, CLEAR employs masked prompts $M^a$ and $M^o$ to perform cloze-style reasoning and infer high-level primitive representations. The re-ranking scores are finally computed by matching the high-level primitives with the primitive sets in a single-modality textual space.
\subsection{The Base Model}
We introduce a classical three-branch (attribute, object, and composition) CZSL framework as the Base Model to perform coarse-grained recognition, where logits are obtained by cross-modal matching between visual representations and primitive representations in each branch.

\noindent\textbf{Visual Representations.} 
Given a batch of images with batch size $B$, we employ the CLIP image encoder $E_v$ equipped with lightweight adapters to extract the global image representation $f_v^{cls} \in \mathbb{R}^{B \times D}$, which is directly used as the image feature $f_v^c$ for the composition branch. To facilitate feature disentanglement, $f_v^c$ is further processed by two independent MLPs, yielding an attribute-oriented image feature $f_v^a$ and an object-oriented image feature $f_v^o$, respectively.

\noindent\textbf{Primitive Representations.}
Inspired by \cite{Troika}, we construct three types of learnable textual prompts for each attribute–object primitive pair $w_i^a \in \mathcal{A}$ and $w_j^o \in \mathcal{O}$, as well as their composition $(w_i^a, w_j^o) \in \mathcal{C}$. Specifically, we define the attribute prompt $\theta_i^a = [p_0^a, \dots, p_m^a, w_i^a]$, the object prompt $\theta_j^o = [p_0^o, \dots, p_m^o, w_j^o]$, and the composition prompt $\theta_{i,j}^c = [p_0^c, \dots, p_m^c, w_i^a, w_j^o]$. Here, $p_{0:m}^a$, $p_{0:m}^o$, and $p_{0:m}^c$ denote learnable prefix tokens, which are initialized with the phrase \textit{``a photo of''}. We employ the CLIP text encoder $E_t$ to encode all primitives, obtaining the corresponding text features $t^a \in \mathbb{R}^{|\mathcal{A}| \times D}$, $t^o \in \mathbb{R}^{|\mathcal{O}| \times D}$, and $t^c\in \mathbb{R}^{|\mathcal{C}| \times D}$.

\noindent\textbf{Three-Branch Basic Logits.} 
Given the visual features and primitive representations from the three branches, for branch $r \in \{a, o, c\}$, the base logit of the image $n$ corresponding to class $k$ with the temperature $\tau$ can be computed as
\begin{equation}
\label{eq:logits}
z^{r}_{n,k} = \frac{ f^{r}_{v,n} \cdot t^{r}_{k} }
{\| f^{r}_{v,n} \|_2 \, \| t^{r}_{k} \|_2 } \cdot \frac{1}{\tau},
\end{equation}

\subsection{Cloze-style Reasoning}
\label{sec:rea}
Given the predictions of the Base Model, we regard the resulting candidate primitive set as having eliminated most instance-irrelevant categories. In an ideal case, it contains multiple sub-primitives of the instance’s ground-truth (GT) primitive, which jointly characterize the GT primitive (e.g., ``tall”, ``wide” and ``long” collectively describe “large”). However, in practice, the candidate set often includes noisy primitives or suffers from missing ones. Therefore, we design a cloze-style reasoning process to abstract the most appropriate primitive text from this noisy candidate set, and it includes two parts:

\subsubsection{Masked Representations}

Serving as carriers of conditional information for reasoning, following the commonly used Masked Language Modeling (MLM) paradigm, we introduce two additional learnable [MASK] tokens $ m^a, m^o \in \mathbb{R}^{1 \times D}$, which are used to denote the attribute and object to be decoded, respectively. They are initialized using the average pooling of the primitive tokens, namely $\bar{w}^a$ and $\bar{w}^o$.

Then, we replace the original attribute and object primitive with the corresponding [MASK] token, yielding two masked prompts $M^a = [p_0^a, \dots, p_m^a, m^a]$ and $M^o = [p_0^o, \dots, p_m^o, m^o]$. Here, the prefix tokens are kept identical to those in the Base Model to ensure that the masked token reside in the same feature space as the corresponding primitives. 

However, unlike MLM, we typically align primitives with images at the sentence level in CZSL. Consequently, recovering the word at the [MASK] position in the same way as MLM thus contradicts the basic training objective. Therefore, we instead use the sentence-level features $\hat{t}^a$ and $\hat{t}^o$ encoded by the CLIP text encoder as the final masked representations for subsequent reasoning, i.e., 

\begin{equation}
\hat{t}^a = E_t(M^a),\quad\hat{t}^o = E_t(M^o)
\end{equation}

\subsubsection{Reasoner}

During reasoning, the masked representations are progressively transformed into high-level primitives, both of which reside in the same CLIP feature space. Therefore, as shown in Fig.~\ref{fig:overview} (b), we implement the Reasoner as a lightweight 6-layer Transformer decoder.

Concretely, taking the attribute branch as an example, given a batch of images of batch size $B$, we first repeat the masked representations $\hat{t}^a$ along the batch dimension as queries. Then, based on the Base Model’s prediction $p(a \mid f_v^{a}) \in \mathbb{R}^{B \times|\mathcal{A}|}$, we select, for each instance, the top-$K$ primitives with the highest predicted probabilities from the attribute primitive set $t^a$, to form a candidate set $t_{Can}^a \in \mathbb{R}^{B \times K \times D}$, i.e.,
\begin{equation}
    Index = TopK\big(p(a \mid f_v^{a})\big), \quad
t_{Can}^a = t^a[Index].
\end{equation}
In addition, we further feed the fine-grained visual features from CLIP, denoted as $f_v^{P} \in \mathbb{R}^{B \times S \times D}$, into the reasoning process, thereby enabling the model to identify the most suitable concrete primitive based on image details when the candidate set is insufficient, where $S$ denotes the number of CLIP patches. As a result, the complete contextual feature $f_{kv} \in \mathbb{R}^{B \times (K+S) \times D}$ is constructed by concatenating the candidate primitives $t_{Can}^a$ and the patch-level visual features:
\begin{equation}
    f_{kv} = Cat(t_{Can}^a; f_v^{P}),
\end{equation}
where $Cat(\cdot)$ denotes the concatenation operation. In practice, $K$ is defined as a fixed proportion of the total number of primitive categories in each branch, and its effect is further analyzed in Sec.~\ref{sec:k}.

\subsection{Re-ranking and Loss}
After the reasoning process, the masked representations $\hat{t}^a$ and $\hat{t}^o$ aggregate visual details and prior semantic knowledge, yielding the primitive variants $\hat{t}^a_{out}$ and $\hat{t}^o_{out}$. Although the reasoning is conditioned on the candidate set, it aims to summarize the most appropriate primitive; therefore, $\hat{t}^a_{out}$ and $\hat{t}^o_{out}$ may exhibit primitive characteristics beyond the candidate set. Therefore, taking the attribute branch as an example, unlike classical re-ranking strategies that only reweight the candidate score, we compute re-ranking scores $\hat{\mathbf{z}}^a \in \mathbb{R}^{B \times |\mathcal{A}|}$ over the entire primitive space, i.e.,

\begin{equation}
\label{eq:logits}
\hat{\mathbf{z}}^a
=
\frac{
\hat{t}^a_{out} \cdot t^a
}{
\left\| \hat{t}^a_{\mathrm{out}} \right\|_2
\left\| t^a \right\|_2
}\cdot \frac{1}{\tau},
\end{equation}
Here, $\tau$ is the temperature parameter. We perform re-ranking directly at the logit level, so that the Base Model is supervised only through the fused results, granting some tolerance to its coarse-grained classification. Let $y_n^a \in \{1,\dots,|\mathcal{A}|\}$ denote the ground-truth attribute label of the $n$-th sample, the final loss is defined using a standard cross-entropy loss:

\begin{equation}
\label{eq:ce_a}
\hat{\mathcal{L}}^a =
-\frac{1}{N} \sum_{n=1}^{N} {p}(y_n^a|n),
\end{equation}
\begin{equation}
    {p}(a_i|n)=\frac{\exp( z^{a}_{n,a_i}+\hat{z}^{a}_{n,a_i})}
{\sum_{k=1}^{|\mathcal{A}|} \exp( z^{a}_{n,k})}.
\end{equation}
The object branch is trained in the same manner, by replacing
$(a, \mathcal{A}, \mathbf{z}^a, \hat{\mathbf{z}}^a)$ with $(o, \mathcal{O}, \mathbf{z}^o, \hat{\mathbf{z}}^o)$. The compositional branch adopts the loss function of the Base Model:
\begin{equation}
\label{eq:ce_a}
\mathcal{L}^c =
-\frac{1}{N} \sum_{n=1}^{N}{p}(y_n^c|n)=-\frac{1}{N} \sum_{n=1}^{N}
\frac{\exp( z^{c}_{n,y_n^a})}
{\sum_{k=1}^{|\mathcal{C}^s|} \exp( z^{c}_{n,k})}.
\end{equation}
The overall learning objective is defined as
\begin{equation}
\label{eq:bas}
\mathcal{L} =
\lambda^a \hat{\mathcal{L}}^a +
\lambda^o \hat{\mathcal{L}}^o +
\lambda^c \mathcal{L}^c ,
\end{equation}
where $\lambda^a = \lambda^o = \lambda^c = 1$, following empirical practice. The final probability for a given sample is calculated as the sum of the attribute, object and composition probability:
\begin{equation}
{p'}(c_{i,j}|x)=p(c_{i,j}|x)+p(a_i|x)+p(o_j|x).
\end{equation}

\begin{table*}[t!]
\centering
\caption{The experimental results for both closed/open-world settings. The best performance are highlighted in bold.}
\resizebox{0.8\linewidth}{!}{
\begin{tabular}{r|c|cccc|cccc|cccc}
\toprule
\multirow{2}{*}{\textbf{Method}} & \multirow{2}{*}{\textbf{Venue}} & \multicolumn{4}{c|}{C-GQA} & \multicolumn{4}{c|}{MIT-States} & \multicolumn{4}{c}{UT-Zappos} \\ \cmidrule{3-14}
                       &  & S    & U    & HM   & AUC    &   S    & U    & HM   & AUC    &   S    & U    & HM   & AUC  \\ \midrule
\multicolumn{14}{c}{Closed-world Results}    \\                                                                               \midrule
PLID\cite{PLID}   & ECCV'24   & 38.8 & 33.0 & 27.9 & 11.0 & 49.7 & 52.4 & 39.0 & 22.1 & 67.3 & 68.8 & 52.4 & 38.7    \\
CDS-CZSL\cite{li2024context}  &  CVPR'24          & 38.3 & 34.2 & 28.1 & 11.1 & 50.3 & 52.9 & 39.2 & 22.4 & 63.9 & 74.8 & 52.7 & 39.5    \\ 
Troika\cite{Troika} &  CVPR'24            & 41.0 & 35.7 & 29.4 & 12.4 & 49.0 & 53.0 & 39.3 & 22.1 & 66.8 & 73.8 & 54.6 & 41.7    \\ 
IMAX\cite{10737702}& TPAMI'25 &39.7 & 35.8& 29.8 & 12.8& 48.7& 53.8  & 39.1 & 21.9  & 69.3 &70.7 & 54.2 & 40.6      \\ 
CLUSPRO\cite{CLUSPRO}   & ICLR'25 &   44.3 & 37.8 & 32.8 & 14.9& \underline{52.1} & \textbf{54.0} & \textbf{40.7} & \underline{23.8}& \textbf{70.7}& 76.0 & \textbf{58.5} & \textbf{46.6} \\
LOGICZSL\cite{logic}   & CVPR'25 &   44.4 & 39.4 & 33.3 & 15.3& 50.8 & \underline{53.9} & 40.5 & 23.4& 69.6 & 74.9 & 57.8 & 45.8 \\
CPF\cite{CPF}& ICCV'25 &\underline{44.8} & \underline{39.6}& \underline{33.6} & \underline{15.4}& 51.6& 53.4  & 40.3 & 23.4 & 69.3 & \underline{76.3} & 57.9 & 45.8       \\ 
\midrule

\multicolumn{2}{c|}{Base Model} &   38.5 & 33.2 & 27.9 & 11.0   &   49.2 & 52.6 & 38.7& 21.8  &   64.4 & 70.7 & 51.9 & 37.8 \\ 
\multicolumn{2}{c|}{\textbf{CLEAR~(Ours)}}  &   \textbf{45.5} & \textbf{41.9} & \textbf{34.9} & \textbf{16.6}   &   \textbf{52.4} & 53.7 & \underline{40.6} & \textbf{24.0}  &   \underline{69.9}&\textbf{76.4} & \underline{58.2} & \underline{46.3} \\

\midrule
\multicolumn{2}{c|}{\color{red}$\Delta$}  &   \color{red}${\uparrow7.0}$ & \color{red}${\uparrow8.7}$ & \color{red}${\uparrow7.0}$ & \color{red}${\uparrow5.6}$   &   \color{red}${\uparrow3.2}$ & \color{red}${\uparrow1.1}$ & \color{red}${\uparrow1.9}$ & \color{red}${\uparrow2.2}$  & \color{red}${\uparrow5.5}$ & \color{red}${\uparrow5.7}$ & \color{red}${\uparrow6.3}$ & \color{red}${\uparrow8.5}$ \\ \midrule

\multicolumn{14}{c}{Open-world Results}                                                                                    \\ \midrule
PLID\cite{PLID}   & ECCV'24   & 39.1 & 7.5 & 10.6 & 2.5 & 49.1 & 18.7 & 20.4 & 7.3& 67.6 & 55.5 & 46.6 & 30.8 \\
CDS-CZSL\cite{li2024context} & CVPR'24      & 37.6 & 8.2 & 11.6 & 2.7 & 49.4 & \underline{21.8} & 22.1 & 8.5& 64.7 & 61.3 & 48.2 & 32.3  \\
Troika\cite{Troika}     &  CVPR'24   & 40.8 & 7.9 & 10.9  & 2.7 & 48.8 & 18.4 & 20.1 & 7.2  &   66.4 & 61.2 & 47.8 & 33.0   \\
IMAX\cite{10737702}                  & TPAMI'25 &38.7 & 7.9& 11.2 & 2.5 & 50.2& 18.6  & 21.4 & 7.6& 68.4&57.3 & 47.5 & 32.3       \\ 
CLUSPRO\cite{CLUSPRO}   & ICLR'25 &   41.6 & 8.3 & 11.6 & 3.0& \underline{51.2}& \textbf{22.1} & \textbf{23.0} &\textbf{9.3}& \textbf{71.0} & \textbf{66.2} & \textbf{54.1} & \textbf{39.5} \\
LOGICZSL\cite{logic}   & CVPR'25 &   43.7 & 9.3 & 12.6 & 3.4& 50.7 & 21.4 & 22.4 & 8.7& \underline{69.6} & 63.7 & 50.8 & 36.2 \\
CPF\cite{CPF}   & ICCV'25 &   \underline{44.5} & \underline{9.3} & \underline{13.0} & \underline{3.6}&51.0 & 20.8 & 22.2 & 8.6& 69.3 & 63.8 & 50.7 & 36.6 \\
\midrule

\multicolumn{2}{c|}{Base Model}&   38.9 & 8.1 & 11.3 & 2.6& 48.5 & 18.0 & 19.8 & 7.0& 66.8 & 59.0 & 47.6 & 32.5 \\
\multicolumn{2}{c|}{\textbf{CLEAR~(Ours)}} &   \textbf{45.8} & \textbf{11.5} & \textbf{15.4} & \textbf{4.6}& \textbf{52.4} & 21.5 & \underline{22.7} &\underline{9.1}& 68.7 & \underline{65.8} & \underline{52.6} & \underline{38.1} \\
\midrule
\multicolumn{2}{c|}{\color{red}$\Delta$}  &   \color{red}${\uparrow6.9}$ & \color{red}${\uparrow3.4}$ & \color{red}${\uparrow4.1}$ & \color{red}${\uparrow2.0}$   &   \color{red}${\uparrow3.9}$ & \color{red}${\uparrow3.5}$ & \color{red}${\uparrow2.9}$ & \color{red}${\uparrow2.1}$  & \color{red}${\uparrow1.9}$ & \color{red}${\uparrow6.8}$ & \color{red}${\uparrow5.0}$ & \color{red}${\uparrow5.6}$ \\
\bottomrule
\end{tabular}}
\label{result}
\end{table*}

\section{Experiment}
\subsection{Experiment Setup}
\noindent\textbf{Datasets and Metrics.} We evaluate CLEAR on three widely used benchmarks for CZSL: UT-Zappos\cite{ut}, MIT-States\cite{mit}, and C-GQA\cite{cgqa}. Following prior works\cite{cgqa}, we adopt the standard dataset splits for fair comparison and report Seen (S) and Unseen (U) composition accuracy, the Harmonic Mean (HM) of Seen and Unseen accuracy, and the Area Under Curve (AUC), with AUC being the most representative metric.


\noindent\textbf{Implementation Details.} CLEAR is trained for 15 epochs with Adam optimizer for all datasets. We jointly train the Base Model and the reasoner rather than in a stage-wise manner, using the pre-trained CLIP ViT-L/14 model\cite{clip} as both the text and image encoder. All training and evaluation are conducted on 2 NVIDIA 4090 GPUs.

\subsection{Main Results}

To intuitively illustrate CLEAR’s capability in addressing the contextual dependency of primitives, we compare CLEAR with state-of-the-art methods that are explicitly designed for the same problem in Table~\ref{result}, including CLUSPRO\cite{CLUSPRO}, LOGICZSL\cite{logic}, and CPF\cite{CPF}. They respectively model primitive contextual dependencies through clustering, the incorporation of external knowledge, and an object-prioritized focus. Note that CPF reports results only on C-GQA, we thus reproduce its performance on the other two datasets.

The results show that CLEAR’s advantage is positively correlated with the scale of the dataset. In particular, on the large-scale C-GQA benchmark, CLEAR significantly and consistently outperforms all other SOTA approaches, demonstrating that our strategy of decoupling conditional information via candidate sets is better suited for complex real-world scenarios. On MIT-States, CLEAR still maintains a performance advantage. However, on the small-scale UT-Zappos dataset, CLEAR outperforms CPF and LOGICZSL, which uses additional knowledge, but only achieves comparable performance to the clustering-based CLUSPRO. This may be attributed to the fact that when the number of primitives is extremely small, modeling-based approaches can easily enumerate and capture almost all the compositional patterns seen.

In addition, we also report the performance of the independently trained Base Model in Table~\ref{result}. Without extra components, its performance is even lower than that of the commonly used baseline Troika\cite{Troika}. CLEAR consistently yields substantial improvements over the Base Model across all three datasets, indicating that re-ranking effectively guides the Base Model to capture complex contextual information.

\subsection{Ablation Study}

\subsubsection{The Top-K in CLEAR}
\label{sec:k}
\begin{figure}[htbp]
\centerline{\includegraphics[width=0.5\linewidth]{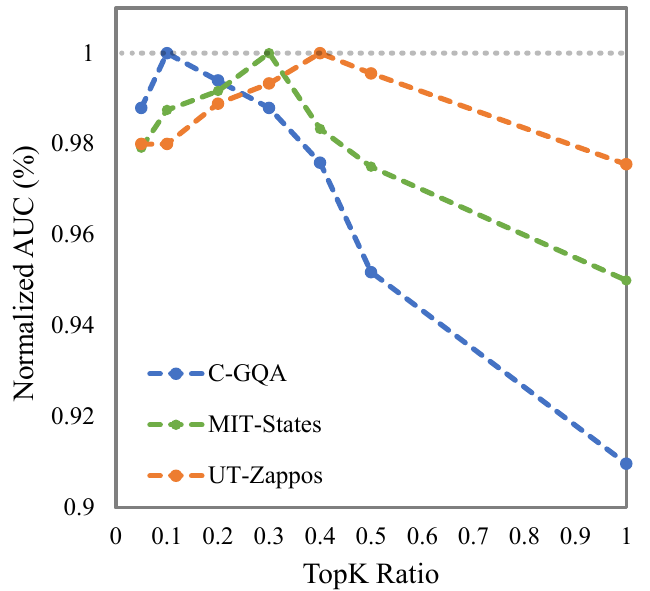}}
\caption{Effect of the TopK ratio $K$ on model performance.
The AUC on each dataset is normalized by its maximum value to facilitate comparison.}
\label{fig:topk}
\end{figure}
We investigate the effect of the TopK parameter $K$ in Fig.~\ref{fig:topk}. In CLEAR, the $K$ for the attribute and object branches are computed by multiplying their respective category cardinalities by a given ratio. We perform a grid search over different ratio on three datasets.

For clarity, we normalize the AUC of each dataset by its own maximum value to highlight relative performance trends. As shown in Fig.~\ref{fig:topk}, increasing the ratio from 0.05 to 1.0 leads to a consistent pattern across all datasets, where performance first improves and then degrades. This observation suggests that K plays a critical role in balancing candidate coverage and noise introduction: a small K tends to miss instance-relevant primitives, whereas an excessively large K introduces many irrelevant candidates. Due to differences in the total number of primitives and the strength of their internal conditional dependencies, the optimal K varies across datasets. The best-performing ratios are 0.1 on C-GQA, 0.3 on MIT-States and 0.4 on UT-Zappos, respectively.

\subsubsection{Reasoning Configuration}

\begin{table}[!ht]
    \centering
    \caption{The ablation study about reasoning configuration.}
\resizebox{0.8\linewidth}{!}{
    \begin{tabular}{c|c|c|c|c}
    \toprule
         \makecell{Candidate\\Context}& \makecell{Visual\\Context} &\makecell{CLIP-Encoded\\Query} & \makecell{Type\\Embedding} & AUC$\uparrow$ \\ \midrule 
        \ding{56} & \ding{56}  & \ding{56} & \ding{56} & 11.0 \\ 
        \ding{51} & \ding{56}  & \ding{56} & \ding{56} & 14.2 \\ 
        \ding{51} & \ding{51}  & \ding{56}   & \ding{56} & 15.6 \\ 
        \ding{51} & \ding{51}  & \ding{51} & \ding{56} & \textbf{16.6} \\ 
        \ding{51} & \ding{51}  & \ding{51} & \ding{51} & 16.1 \\ 
        \bottomrule
    \end{tabular}}
\label{tab:rea}
\end{table}
We evaluate the reasoning configuration on the C-GQA dataset in Table~\ref{tab:rea}, with all experiments performed under the optimal $K$. Here, the candidate and visual contexts correspond to different components of $f_{kv}$. The ``\ding{56}” in ``CLIP-Encoded Query” indicates that the masked representations $\hat{t}^a$ are replaced with directly learnable parameters. ``Type Embedding” is a learnable embedding added to $f_{kv}$ to indicate the positions of different modalities.

As shown in Table~\ref{tab:rea}, introducing candidate context yields a substantial improvement over the Base Model (11.0$\rightarrow$14.2 AUC), indicating that candidate primitives provide a critical semantic prior. Further incorporating visual context leads to additional gains, suggesting that patch-level visual details offer necessary information. The best performance is achieved when CLIP-encoded queries are used together with both candidate and visual contexts (16.6 AUC). This comparison highlights the importance of projecting masked prompts into the same textual feature space as primitives, which helps maintain representation consistency during reasoning. In contrast, adding type embeddings slightly degrades performance, which can be attributed to the interference introduced by explicit modality cues in the highly aligned CLIP feature space.

\begin{figure}[htbp]
\centerline{\includegraphics[width=0.8\linewidth]{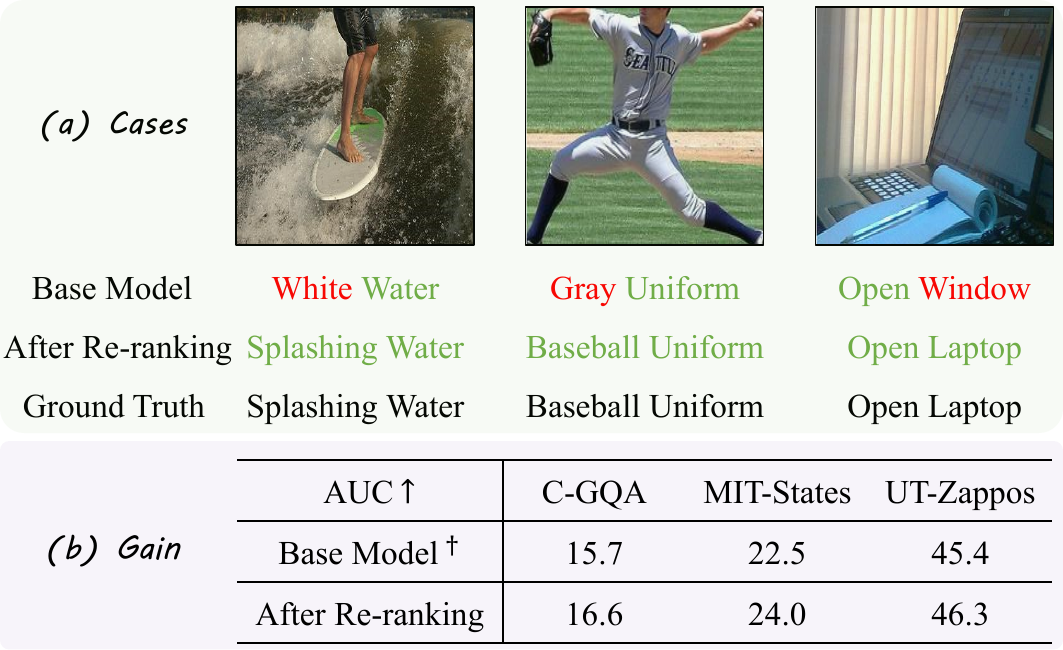}}
\caption{Re-ranking effects of CLEAR. (a) Representative cases comparing the Base Model outputs before re-ranking with the final predictions. (b) Quantitative AUC gains of re-ranking across three datasets.}
\label{fig:vis}
\end{figure}
\subsubsection{Re-ranking and Visualization} Fig.~\ref{fig:vis} provides an analysis of the proposed re-ranking strategy. In this figure, different from Table~\ref{result}, the reported Base Model$^\dagger$ here refers to the predictions of the Base Model after joint training with CLEAR, but before applying the re-ranking module. As shown in Fig.~\ref{fig:vis} (a), the Base Model can be misled by salient but concrete visual cues, resulting in incorrect predictions. In contrast, CLEAR effectively corrects these errors by leveraging conditional reasoning to infer high-level semantics. Fig.~\ref{fig:vis} (b) further reports the quantitative improvements brought by re-ranking on three datasets. The consistent AUC gains demonstrate that CLEAR does not merely enhance the knowledge of the Base Model but exhibits an independent error-correction ability.

\section{Conclusion}
In this paper, we revisit CZSL from the perspective that contextual variations emerge from instance-specific visual cues rather than fixed conditional forms. Based on this insight, we propose CLEAR, a coarse-to-fine framework that performs cloze-style reasoning over candidate primitives and re-ranks base predictions to mitigate biases toward concrete semantics. By reasoning from plausible hypotheses instead of explicitly enumerating primitive variants, CLEAR achieves more flexible and robust compositional generalization.

\section*{Acknowledgment}

This work was supported by the Beijing Advanced Innovation Center
for Future Blockchain and Privacy Computing.

\bibliographystyle{IEEEtran}
\bibliography{icme2026references}

\end{document}